\documentclass[11pt,a4paper]{article}
\usepackage[hyperref]{style/acl2020}
\usepackage{times}
\usepackage{latexsym}
\usepackage{url}
\usepackage{graphicx}
\usepackage{amsmath}
 \usepackage{booktabs}
\usepackage{framed}
\usepackage[titletoc]{appendix}
\usepackage{microtype}

\aclfinalcopy 

\newcommand{\name} {\textsc{splash}}

\title{Speak to your Parser: \\Interactive Text-to-SQL with Natural Language Feedback}

\author{Ahmed Elgohary\thanks{\hspace{1.5mm}Most work was done while the first author was an intern at Microsoft Research.} \\
  University of Maryland, College Park \\
  \texttt{elgohary@cs.umd.edu} \\\And
  Saghar Hosseini, Ahmed Hassan Awadallah \\
  Microsoft Research, Redmond, WA \\
  \texttt{\{sahoss,hassanam\}@microsoft.com} \\}

\begin{document}
\maketitle
\begin{abstract}
We study the task of semantic parse correction with
natural language feedback. Given a natural language utterance, most semantic parsing systems pose the problem as one-shot translation where the utterance is mapped to a corresponding logical form. In this paper, we investigate a more interactive scenario where humans can further interact with the system by providing free-form natural language feedback to correct the system when it generates an inaccurate interpretation of an initial utterance.
We focus on natural language to SQL systems and construct, \name, a dataset of utterances, incorrect SQL interpretations and the corresponding natural language feedback. 
We compare various reference models for the correction task and 
show that incorporating such a rich form of feedback can significantly improve the overall semantic parsing accuracy while retaining the flexibility of natural language interaction. 
While we estimated human correction accuracy is 
81.5\%, our best model achieves 
only 25.1\%,  which leaves a large gap for improvement in future research. 
\name~is publicly available at ~\url{https://aka.ms/Splash_dataset}.
\end{abstract}

\section{Introduction\label{sec:intro}}

Natural language interfaces (NLIs) have been the ``holy grail" of natural language understating and human-computer interaction for decades~\cite{woods1972lunar,codd1974seven,hendrix1978developing,zettlemoyer2005improving}. However, early attempts in building NLIs to databases did not achieve the expected success due to limitations in language understanding capability, among other reasons~\cite{androutsopoulos1995natural, jones1995evaluating}. NLIs have been receiving increasing attention recently motivated by interest in developing virtual assistants, dialogue systems, and semantic parsing systems. NLIs to databases were at the forefront of this wave with several studies focusing on parsing natural language utterances into an executable SQL queries~(Text-to-SQL parsing).

\begin{figure}[t]
  \centering
  \includegraphics[width=0.9\columnwidth]{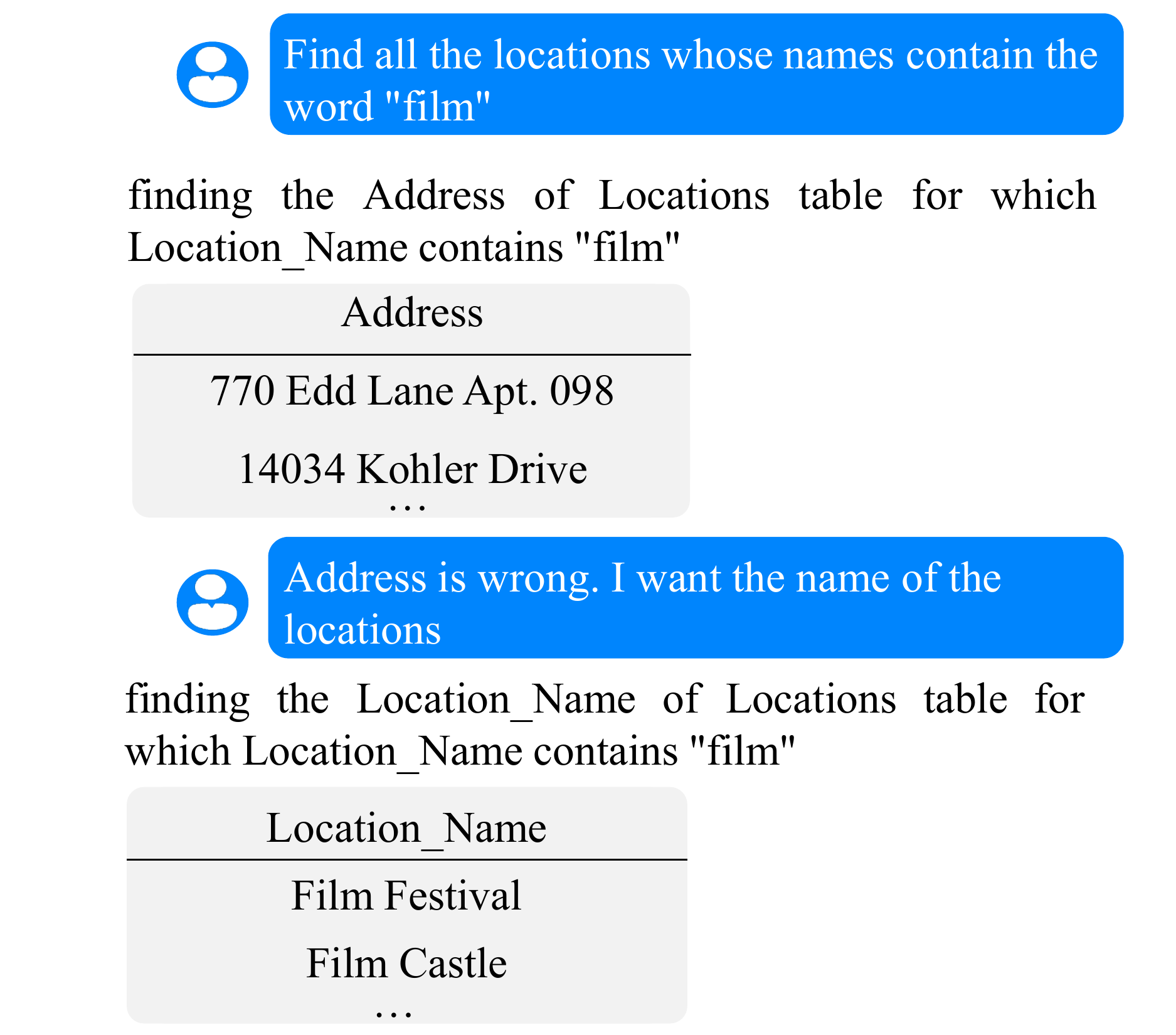}
    \caption{An example of human interaction with a Text-to-SQL system 
            to correct the interpretation of an input utterance.
    The system generates an initial SQL parse, explains it 
    in natural language, and displays the execution result.
    Then, the system uses the human-provided natural language feedback
    to correct the initial parse.}    
  \label{fig:example}
  \vspace{-0.4cm}
 \end{figure}

Most of the work addressing the Text-to-SQL problem (and semantic parsing in general) frames it as a one-shot mapping problem.
We establish~(Section~\ref{sec:erranalysis})~that the majority of parsing
mistakes that recent neural text-to-SQL parsers make are minor. Hence,
it is often feasible for humans to detect and suggest fixes
for such mistakes. ~\newcite{Su:2018:NLI:3209978.3210013} 
make a similar observation about parsing text to API calls~\cite{su2017building} and show that parsing mistakes could be easily corrected if 
humans are afforded a means of providing precise feedback. 
Likewise, an input utterance might be under- or mis-specified, thus extra interactions may be required to generate the desired output similarly to query refinements in information retrieval systems~\cite{dang2010query}.

Humans have the ability to learn new concepts or correct others based on natural language description or feedback. Similarly, previous work has explored how machines can learn from language in tasks such as playing games~\cite{branavan2012learning}, robot navigation~\cite{karamcheti2017tale}, concept learning (e.g., shape, size, etc.) classifiers~\cite{srivastava-etal-2018-zero}, etc. Figure~\ref{fig:example} shows an example of a text-to-SQL system that offers humans the affordance to provide feedback in natural language when the system misinterprets an input utterance. To enable this type of interactions, the system needs to: (1) provide an \emph{explanation} of the underlying generated SQL, (2) provide a means for humans to provide feedback and (3) use the feedback, along with the original question, to come up with a more accurate interpretation.

In this work, we study the task of SQL parse correction with natural language feedback to enable text-to-SQL systems to seek and leverage human feedback to further improve the overall performance and user experience. 
Towards that goal, we make the following contributions: (1) we define the task of SQL parse correction with natural language feedback;
(2) We create a framework for explaining SQL parse in natural language to allow
text-to-SQL  users (who may have a good idea of what kind of information resides on their databases but are not proficient in SQL~\citet{hendrix1978developing}) to provide feedback to correct inaccurate SQL parses; (3) we construct~\name---\textbf{S}emantic \textbf{P}arsing with \textbf{L}anguage \textbf{As}sistance from \textbf{H}umans---a new dataset
of natural language questions 
that a recent neural text-to-SQL parser failed to generate correct interpretation for together with corresponding 
human-provided natural language feedback describing how the 
interpretation should be corrected; and (4) we establish 
several baseline models for the correction task and 
show that the task is challenging for state-of-the-art 
semantic parsing models.

\section{Task\label{sec:task}}
\begin{figure}[t]
  \centering
  \includegraphics[width=\columnwidth]{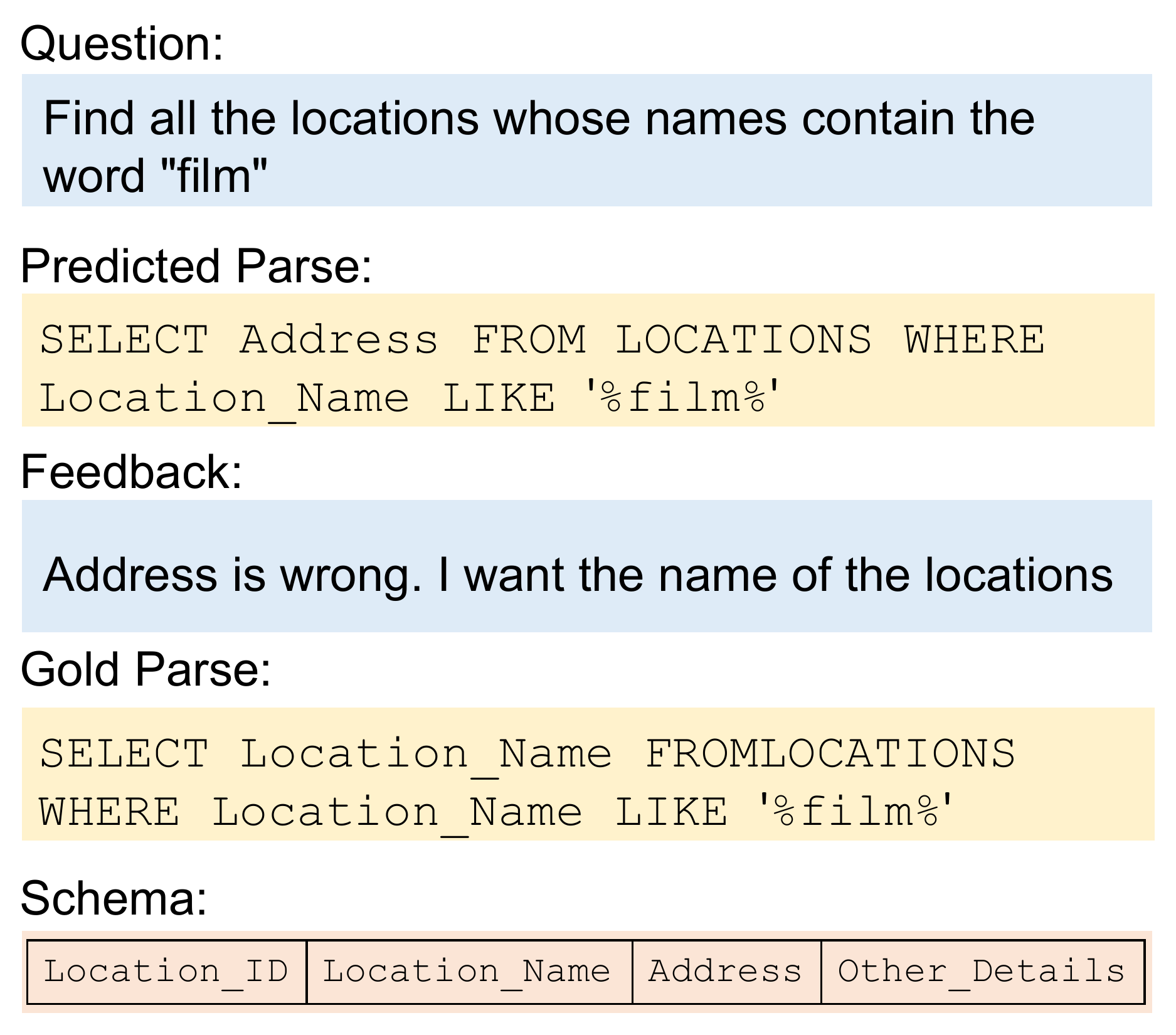}
  \caption{An example from our SQL parse correction
  task (DB Name: \texttt{cre\_Theme\_park} and Table Name: \texttt{Locations}). Given a question, initial predicted parse and
  natural language feedback on the predicted parse, the task is to predict a corrected parse that matches
  the gold parse.}    
\label{fig:dataset_Example}
 \end{figure}
 
We formally define the task of SQL parse correction with natural language feedback.
Given a question $q$, a database schema $s$, a mispredicted parse $p^\prime$,
a natural language feedback $f$ on $p^\prime$, the task is to generate a corrected parse $p$~(Figure~\ref{fig:dataset_Example}).
Following~\newcite{yu-etal-2018-spider}, $s$ is defined as the set of tables, columns in each table and the primary and foreign keys of each table.

Models are trained with tuples
$q$, $s$, $p^\prime$, $f$ and gold parse $p$. At evaluation time, 
a model takes as input tuples in the form $q$, $s$, $p^\prime$, $f$
and hypothesizes a corrected parse $\hat{p}$. We compare $\hat{p}$
and the gold parse $p$ in terms of their exact set match~\cite{yu-etal-2018-spider} 
and report the average matching accuracy across the testing examples as
the model's correction accuracy.

\section{Dataset Construction\label{sec:data_collection}}
In this section, we describe our approach for collecting training data for the SQL parse correction task. We first generate pairs of natural language utterances and the corresponding erroneous SQL parses~(Section \ref{sec:generating-pairs}). 
We then employ crowd workers (with no SQL knowledge) to provide feedback, in natural language, to correct the erroneous SQL (Section~\ref{sec:collection}). 
To enable such workers to provide feedback, we show them an \emph{explanation} of the generated SQL in natural language~
(Section~\ref{sec:explanations}). Finally, to improve the diversity of the natural language feedback, we ask a different set of annotators to paraphrase each feedback. We describe the process in detail in the remainder of this section.

\begin{figure}[t]
  \centering
  \includegraphics[width=\columnwidth]{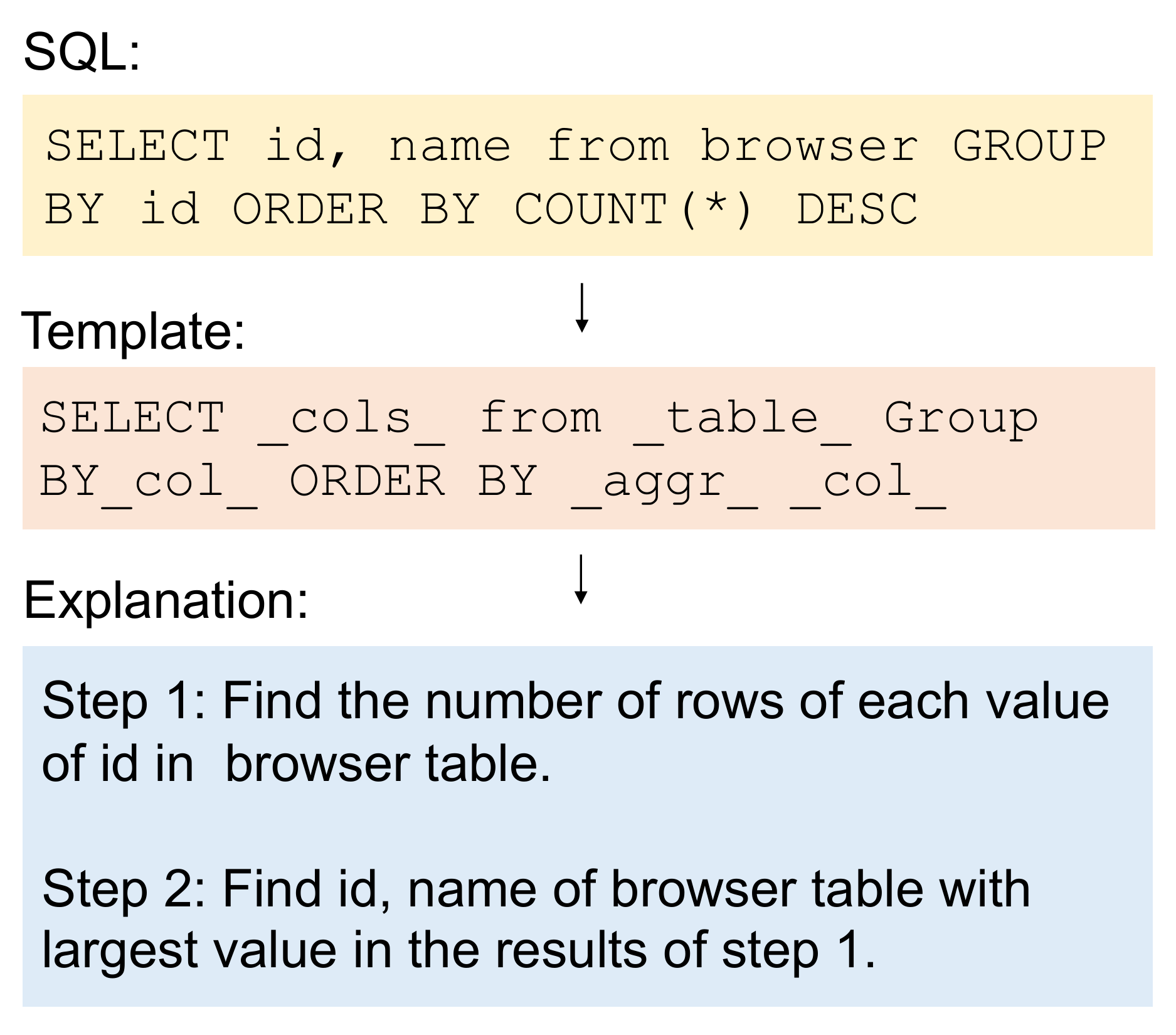}
    \caption{An example of a SQL query, the corresponding template and the generated explanation.}
  \label{fig:explanation_example}
 \end{figure}

\subsection{Generating Questions and Incorrect SQL Pairs}
\label{sec:generating-pairs}
We use the Spider dataset~\cite{yu-etal-2018-spider} as our source of  questions. Spider has several advantages over other datasets. Compared 
to ATIS~\cite{price-1990-evaluation}
and GeoQuery~\cite{Zelle1996LearningTP}, Spider is much larger in scale (200 databases vs. one database, and thousands vs. hundreds of SQL parses). Compared to WikiSQL~\cite{zhong2018seqsql}, 
Spider questions require inducing parses of complex structures
(requiring multiple tables, joining, nesting, etc.).
Spider also adopts a cross-domain evaluation setup in which
databases used at testing time are never seen at training time.

To generate erroneous SQL interpretations of questions in Spider, we opted for using the output of a text-to-SQL parser to ensure that our dataset
reflect the actual distribution of errors that contemporary 
parsers make.
This is a more realistic setup than artificially infusing errors in the gold SQL.
We use the Seq2Struct parser~\cite{shin2019encoding}\footnote{\url{https://github.com/rshin/seq2struct}} to generate erroneous SQL interpretations. Seq2Struct combines
grammar-based decoder of ~\newcite{yin17acl} with
a self-attention-based schema encoding and it reaches 
a parsing accuracy of 42.94\% on the development set of Spider.\footnote{When we started the dataset construction at the beginning of June 2019, we were able to achieve a parsing
accuracy of 41.30\% on Spider's development set which
was the state-of-the-art accuracy at the time.
It is worth noting that unlike current state-of-the-art models, 
Seq2Struct does not use pre-trained language models. It was further developed into 
a new model called RAT-SQL~\cite{rat-sql-2020} which achieved a new state-of-the-art accuracy as of April 2020.}

Note that we make no explicit dependencies on the model used for this step and hence other models could be used as well~(Section~\ref{sec:modelanalysis}).

We train Seq2Struct on 80\% of Spider's training set and apply it to the remaining 20\%, keeping only cases where the generated parses do not 
match the gold parse~(we use the exact set match of~\newcite{yu-etal-2018-spider}). Following the 
by-database splitting scheme of Spider, we repeat the 80-20 training
and evaluation process for three times with different examples in
the evaluation set at each run.
 This results in 3,183 pairs of questions and an erroneous SQL interpretation. To further increase the size of the dataset, we also 
ignore the top prediction in the decoder beam\footnote{We used a beam of size 20.}
and use the following predictions. We only include cases where the difference in probability between the top and second to top SQLs is below a threshold of 0.2. The intuition here is that those are predictions that the model was about to make and hence represent errors that the model could have made. That adds 1,192 pairs to our dataset.

\subsection{Explaining SQL\label{sec:explanations}}

In one of the earliest work on natural language interfaces to databases, ~\newcite{hendrix1978developing} note that many business executives, government official and other decision makers  have a good idea of what kind of information residing on their databases. Yet to obtain an answer to a particular question, they cannot use the database themselves and instead need to employ the help of someone who can. As such, in order to support an interactive Text-to-SQL system,  
we need to be able to \emph{explain} the incorrect generated SQL in a way that
humans who are not proficient in SQL can understand.

We take a template-based approach to explain SQL queries in natural language.
We define a template as follows: 
Given a SQL query, we replace literals, table and columns names and
aggregation and comparison operations with
generic placeholders. We also assume that all
joins are inner joins (true for all Spider queries) whose join conditions 
are based on primary and foreign key equivalence (true for more than 96\% of Spider queries).
A query that consists of two subqueries combined with an intersection, union or except
operations is split into two templates that are processed independently and we add
an intersection/union/except part to the explanation at the end. We apply the same
process to the limit operation---generate an explanation of the query without limit, then add a limit-related step at the end. 

We select the most frequent 57 templates used in Spider training set
which cover 85\% of Spider queries.
For each SQL template, we wrote down a corresponding 
explanation template in the form
of steps (e.g., join step, aggregation step, selection step) 
that we populate for each query.
Figure~\ref{fig:explanation_example} shows an example of a SQL queries, its corresponding template
and generated explanations. We also implemented a set of 
rules for compressing the steps based on SQL semantics.
For instance, an ordering step following by a ``limit 1''
is replaced with ``find largest/smallest'' where
``largest'' or ``smallest'' is decided according to
the ordering direction. 

\subsection{Crowdsourcing Feedback}
\label{sec:collection}

We use an internal crowd-sourcing platform similar to Amazon Mechanical Turk to recruit annotators. Annotators are 
only selected based on their performance on other crowd-sourcing tasks and 
command of English. Before working on the task,  annotators go through a brief set of guidelines explaining the task.\footnote{We provide the data collection instructions and a screenshot of the data collection interface in the appendix.} 
We collect the dataset in batches of around 500 examples each.
After each batch is completed, we manually review
a sample of the examples submitted by each annotator
and exclude those who do not provide accurate inputs
from the annotators pool and redo all their 
annotations.

Annotators are shown the original question, the explanation of the generated SQL and asked to: (1) decide whether the generated SQL satisfies the information need in the question and (2) if not, then provide feedback in natural language. The first step is necessary since it helps identifying false negative parses~
(e.g., another correct parse that does not match the gold parse provided in Spider). We also use the annotations of that step to
assess the extent to which our interface enables target users
to interact with the underlying system.
As per our assumption that target users are familiar with the kind of information that is in the database ~\cite{hendrix1978developing},
we show the annotators an overview of the tables in the  database corresponding to the question that includes the table and column names together with examples (first 2 rows) of the content.
We limit the maximum feedback length to 15 tokens
to encourage annotators to provide a correcting
feedback based on the initial parse
(that focuses on the edit to be made) rather
than describing how the question should be answered.

A total of 10 annotators participated in this task. They were compensated based on an hourly rate (as opposed to per annotation) to encourage them to optimize for quality and not quantity. They took an average of 6 minutes per annotation. 

To improve the diversity of the feedback we collect, we ask a separate set of annotators to generate a paraphrase of each feedback utterance. 
We follow the same annotators quality control measures as in the
feedback collection task. An example instance from the dataset is shown in Figure~\ref{fig:dataset_Example}.

\subsection{Dataset Summary}

\begin{table}[t!]
\small
\centering
\begin{tabular}{llll}
\toprule
\textbf{Number of} & \textbf{Train} & \textbf{Dev} & \textbf{Test}\\
\hline
Examples & 7,481 & 871 & 962\\
Databases & 111 & 9 & 20\\
Uniq. Questions & 2,775 & 290 & 506\\
Uniq. Wrong Parses & 2,840 & 383 & 325\\
Uniq. Gold Parses & 1,781 & 305 & 194\\
Uniq. Feedbacks & 7,350 & 860 & 948\\
Feedback tokens (Avg.) & 13.9 & 13.8 & 13.1\\
 \bottomrule
\end{tabular}
\caption{\name~summary}
\label{tab:datastats}
\end{table}

Overall, we ask the annotators to annotate 5409 example (427 of which had the correct SQL parse and the remaining had an incorrect SQL parse). Examples with correct parse are included to test whether the annotators are able to identify correct SQL parses given their explanation and the question. Annotators are able to identify the correct parses as correct 96.4\% of the time. For the examples whose predicted SQL did not match the gold SQL, annotators still marked 279 of them as correct. Upon manual examinations, we found that annotators were indeed correct in doing so 95.5\% of the time. Even though the predicted and gold SQLs did not match exactly, they were equivalent (e.g.,  \texttt{'price between 10 and 20'} vs. \texttt{'price $\geq$ 10 and price $\leq$ 20'}). 

After paraphrasing, we ended up with 9,314 question-feedback pairs,
8352 of which correspond to questions in the Spider training split and 962 from the spider development split. We use all the data from the Spider development split as our test data. We hold out 10\% of the remaining data (split by database) to use as our development set and use the rest as the training set.
Table~\ref{tab:datastats} provides a summary of 
the final dataset.

\section{Dataset Analysis\label{sec:data_analysis}}
We conduct a more thorough analysis of \name~in this section. 
We study the characteristics of the mistakes made by the parser as well as characteristics of the natural language feedback.

\subsection{Error Characteristics\label{sec:erranalysis}}
\begin{figure}[t]
  \centering
  \includegraphics[width=0.9\columnwidth, viewport = 40 100 780 520,clip]{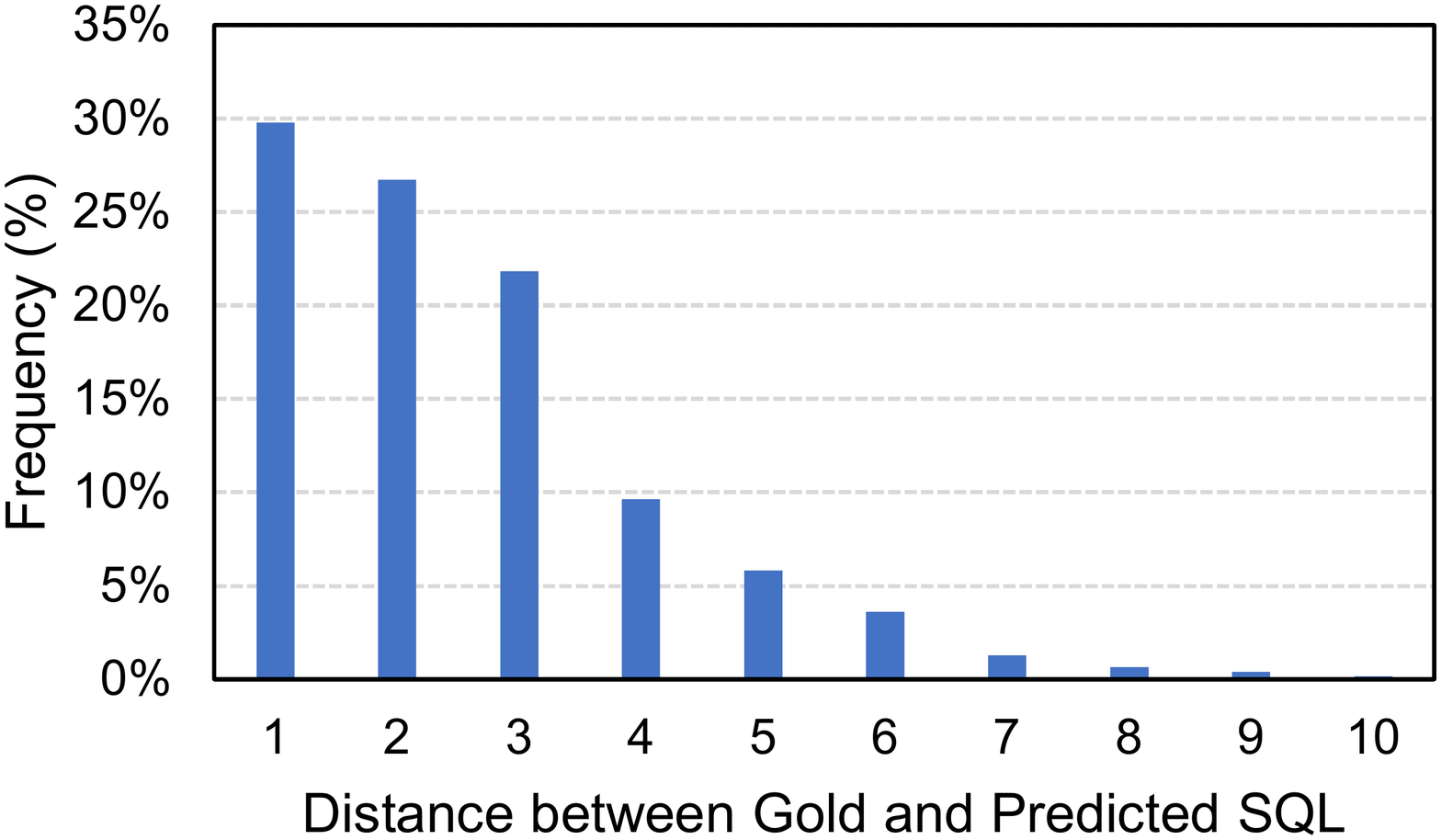}
    \caption{A histogram of the distance between the gold and the predicted SQL.}
  \label{fig:distvsfreq}
 \end{figure}

We start by characterizing the nature of errors usually made by the models in parsing the original utterance to SQL. To understand the relation between the gold and the predicted SQL, we measure the edit distance between them for all cases for which the model made a mistake in the SQL prediction. We measure the edit distance by
the number of edit segments (delete, insert, replace) between
both parses. We find the minimal sequence of token-level
edits using the levenshtein distance algorithm. Then,
we combine edits of the same type (delete, insert, replace) 
applied to consecutive positions in the predicted parse in 
 one segment.
Figure~\ref{fig:distvsfreq} shows a frequency histogram of different values of edit distance. We observe that most inaccurate predictions lie within a short distance from the correct SQL ($78\%+$ within a distance of $3$ or less).

In addition to the number of edits, we also characterize the types of edits needed to convert the predicted SQL to the gold one. Our edit distance calculations support three operations replace, insert and delete. Those correspond to 58\%,, 31\% and 11\% of the edit operations respectively. Most of the edits are rather simple and require replacing, inserting or deleting a single token (68.1\% of the edits). The vast majority of those correspond to editing a schema item (table or column name): 89.2\%, a SQL keyword (e.g., order direction, aggregation, count, distinct, etc.): 7.4\%, an operator (greater than, less than, etc.): 2.2\% or a number (e.g. for a limit operator): 1.2\%.

\begin{figure}[t]
  \centering
  \includegraphics[width=0.9\columnwidth, viewport = 60 85 780 538,clip]{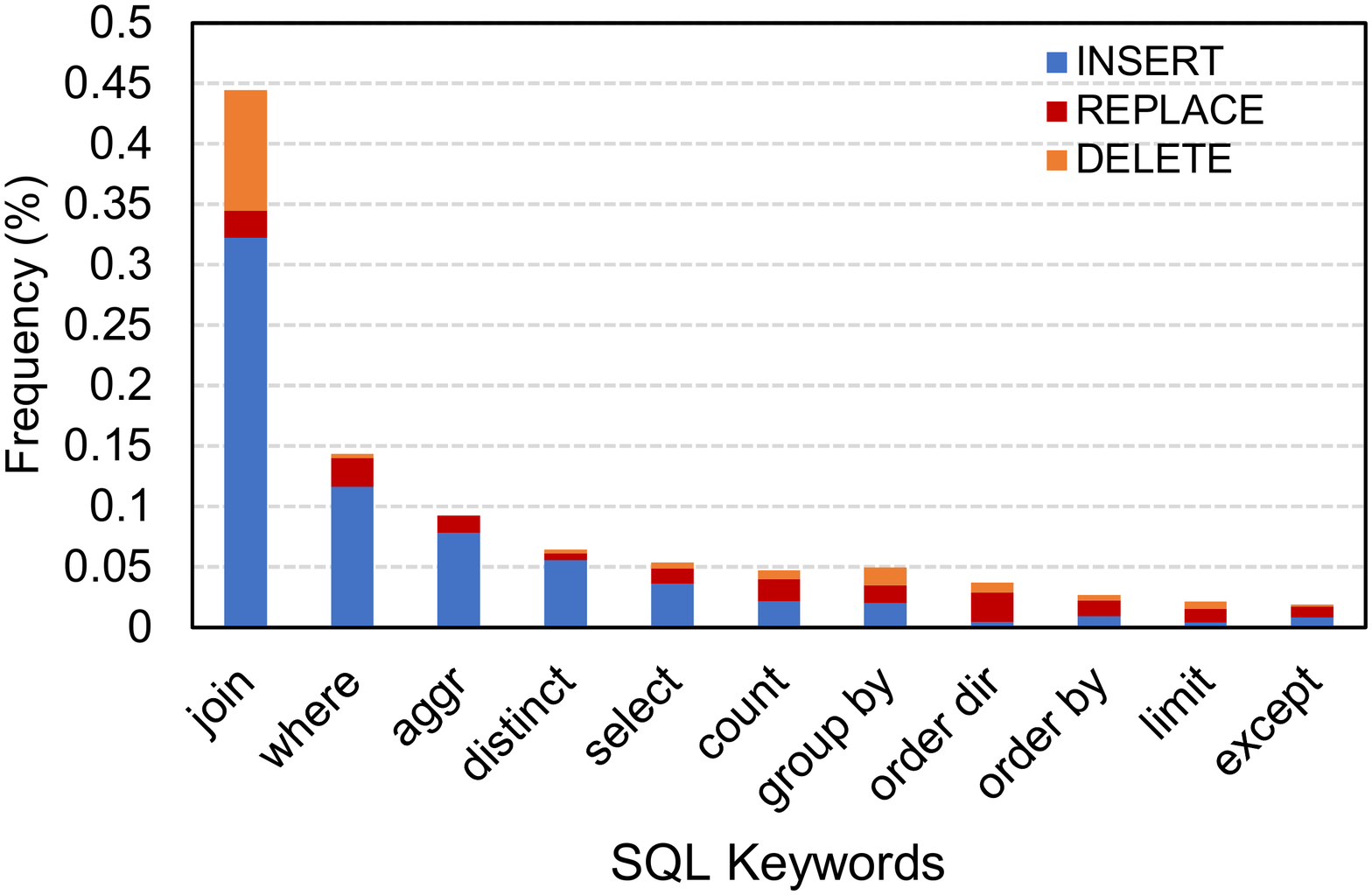}
    \caption{A histogram of different SQL keywords appearing in edits (between the gold and predicted SQL) and their distribution across edit types (replace, insert or delete).}
  \label{fig:errokwfreq}
 \end{figure}

The edits between the predicted and generated SQL spanned multiple SQL keywords. The distribution of different SQL keywords appearing in edits and their distribution across edit types (replace, insert or delete) is shown in Figure~\ref{fig:errokwfreq}. Note that a single edit could involve multiple keywords (e.g., multiple joins, a join and a where clause, etc.). Interestingly, many of the edits involve a \emph{join} highlighting that handling utterances that require a join is harder and more error prone. Following \emph{join}, most edits involve \emph{where} clauses (making the query more or less specific), aggregation operators, counting and selecting unique values. 

The error analysis demonstrates that many of the errors made by the model are in fact not significant and hence it is reasonable to seek human feedback to correct them.

\subsection{Feedback Characteristics}
\label{sec:feedback_characteristics}

\begin{table*}[ht]
\small
\centering
\begin{tabular}{ll}
\toprule
 \multicolumn{2}{l}{\textbf{Complete Feedback: [81.5\%]} }  \\
 Question: & Show the types of schools that have two schools. \\
 Pred. SQL: & \texttt{SELECT TYPE FROM school GROUP BY TYPE HAVING count(*) >= 2}\\
 Feedback: & You should not use greater than. \\ 
\midrule
\multicolumn{2}{l}{\textbf{Partial Feedback: [13.5\%]} }  \\
 Question: & What are the names of all races held between 2009 and 2011? \\
 Pred. SQL: & \texttt{SELECT country FROM circuits WHERE lat BETWEEN 2009 AND 2011}\\
 Feedback: & You should use races table. \\ 
 \midrule
\multicolumn{2}{l}{\textbf{Paraphrase Feedback: [5.0\%]} }  \\
 Question: & What zip codes have a station with a max temperature greater than or equal to 80\\
 & and when did it reach that temperature? \\
 Pred. SQL: & \texttt{SELECT zip\_code FROM weather WHERE min\_temperature\_f} \\
 & \texttt{> 80 OR min\_sea\_level\_pressure\_inches > 80}\\
 Feedback: & Find date , zip code whose max temperature f greater than or equals 80. \\ 
 \bottomrule
\end{tabular}
\caption{Examples (question, predicted SQL and feedback) of complete, partial and paraphrase feedback}
\label{tab:feedback-type2-examples}
\vspace{-0.1cm}
\end{table*}

To better understand the different types of feedback our annotators provided, we sample 200 examples from the dataset, and annotate them with the type of the feedback. We start by assigning the feedback to one of three categories: (1) Complete: the feedback fully describes how the predicted SQL can be corrected , (2) Partial: the feedback describes a way to correct the predicted SQL but only partially and (3) Paraphrase: the feedback is a paraphrase of the original question. The sample had 81.5\% for Complete, 13.5\% for Partial and 5.0\% for Paraphrase feedback. Examples of each type of feedback are shown in Table~\ref{tab:feedback-type2-examples}. Upon further inspection of the partial and paraphrase feedback, we observe that they mostly happen when the distance between the predicted and gold SQL is high (major parsing errors). As such, annotators opt for providing partial feedback (that would at least correct some of the mistakes) or decide to rewrite the question in a different way. 

We also annotate and present the types of feedback, in terms of changes the feedback is suggesting, in Table~\ref{tab:feedback-type-examples}. Note that the same feedback may suggest multiple changes at the same time. The table shows that the feedback covers a broad range of types, which matches our initial analysis of error types. We find that a majority of feedback is referencing the retrieved information. In many such cases, the correct information has not been retrieved because the corresponding table was not used in the query. This typically corresponds to a missing inner one-to-one join operation and agrees with our earlier analysis on edit distance between the gold and predicted SQL. The second most popular category is incorrect conditions or filters followed by aggregation and ordering errors. We split the first two categories by whether the information/conditions are missing, need to be replaced or need to be removed. We observe that most of the time the information or condition needs to be replaced. This is followed by missing information that needs to be inserted and then unnecessary ones that need to be removed.

We heuristically identify feedback patterns for each collected feedback. To identify the feedback pattern, we first locate the central predicate in the feedback sentence using a semantic role labeler~\cite{he-etal-2015-question}.
Since some feedback sentences can be broken into multiple sentence fragments, a single feedback may contain more than one central predicate. For each predicate, we identify its main arguments. We represent every argument with its first non stopword token. To reduce the vocabulary, we heuristically identify column mentions and replace them with the token 'item'.

\begin{figure}[t]
\vspace{-0.8cm}
  \centering
  \includegraphics[width=\columnwidth, viewport = 140 60 650 630,clip]{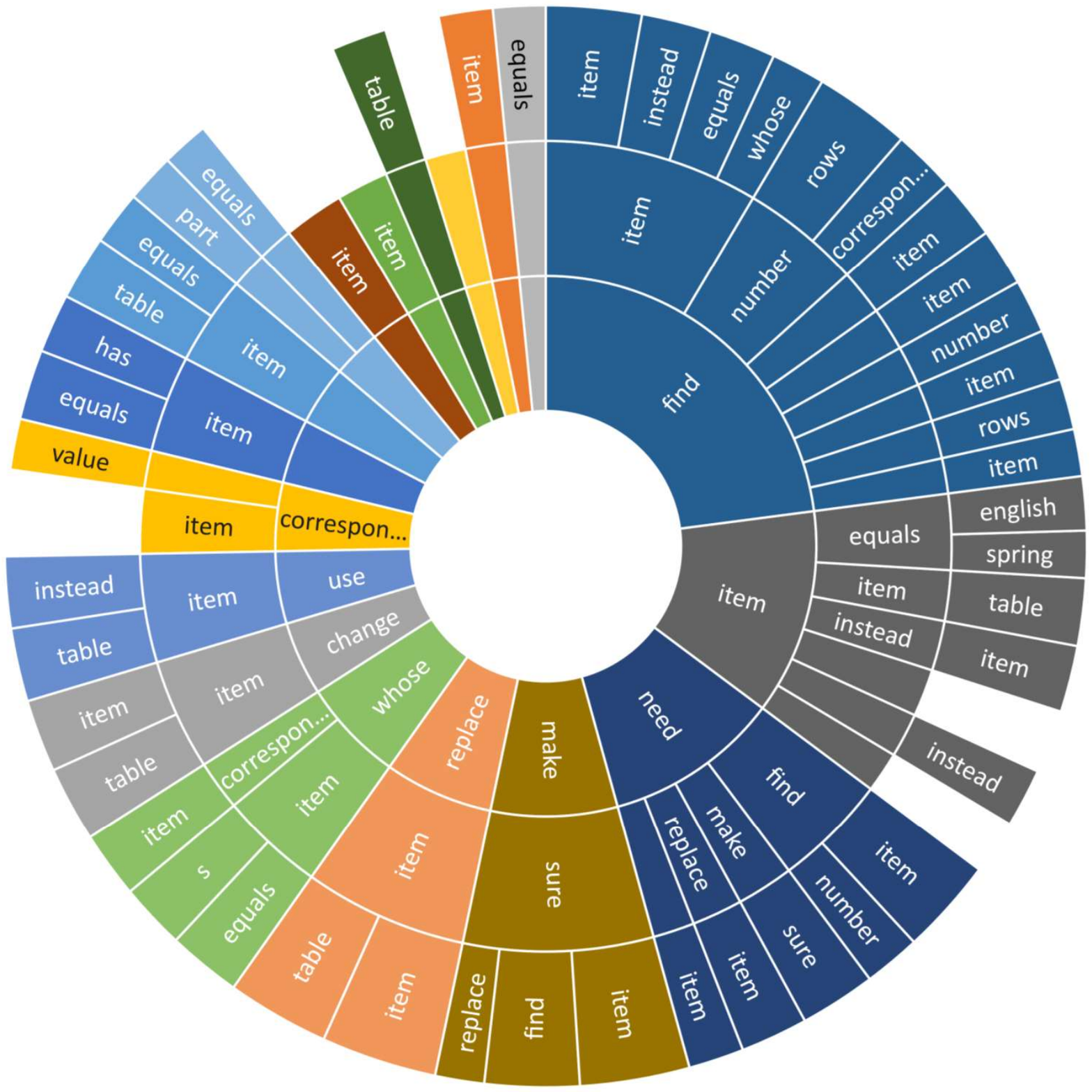}
    \caption{Patterns of feedback covered in our dataset. Patterns are extracted heuristically using predicates and arguments extracted from the feedback sentence. The figure shows the top 60 frequent patterns.}
  \label{fig:feedbackdiv}
  \vspace{-0.5cm}
 \end{figure}

We visualize the distribution of feedback patterns for the top 60 most frequent patterns in  Figure~\ref{fig:feedbackdiv} , and label the ones shared among multiple patterns. As is shown, our dataset covers a diverse variety of feedback patterns centered around key operations to edit the predicted SQL that reference operations, column names and values.

\begin{table*}[ht!]
\centering
\begin{tabular}{lll}
\toprule
 Feedback Type & \% & Example  \\ \midrule
 Information &  &    \\
 \ \ - Missing & 13\% & I also need the number of different services \\
 \ \ - Wrong & 36\% &  Return capacity in place of height \\
 \ \ - Unnecessary & 4\% & No need to return email address \\
  Conditions &  &    \\
 \ \ - Missing & 10\% & ensure they are FDA approved \\
 \ \ - Wrong & 19\% & need to filter on open year not register year \\
 \ \ - Unnecessary & 7\% & return results for all majors\\
 Aggregation & 6\% & I wanted the smallest ones not the largest\\
 Order/Uniq & 5\%& only return unique values\\
  \bottomrule
\end{tabular}
\caption{Examples of feedback annotators provided for different types\label{tab:feedback-type-examples}}
\vspace{-0.6cm}
\end{table*}

\section{Related Work\label{sec:related_work}}
Our work is linked to multiple  existing research lines including semantic parsing, 
learning through interaction~\cite[inter alia]{li2017dialogue,hancock-etal-2019-learning,Li2017LearningTD}
and learning from natural language supervision~\cite[inter alia]{srivastava-etal-2017-joint,co-reyes2018metalearning,srivastava-etal-2018-zero,hancock-etal-2018-training,Ling17Teaching}.
We discuss connections to the most relevant works.

\textbf{Text-to-SQL Parsing:} Natural language to SQL (natural language interfaces to databases) has been an active field of study for several decades~\cite{woods1972lunar,hendrix1978developing,warren-pereira-1982-efficient, Popescu:2003:TTN:604045.604070,li2014constructing}. This line of work has been receiving increased attention recently driven, in part, by the development of new large scale datasets such as WikiSQL~\cite{zhong2018seqsql} and Spider~\cite{yu-etal-2018-spider}. The majority of this work has focused on mapping a single query to the corresponding SQL with the exception of a few datasets, e.g.,
SParC~\cite{yu2019sparc} and CoSQL~\cite{yu2019cosql},~
that target inducing SQL parses for sequentially related 
questions. While these datasets focus on modeling conversational
dependencies between questions,~\name~evaluates the extent to
which models can interpret and apply feedback on the generated
parses. We empirically confirm that distinction 
in Section~\ref{sec:modelanalysis}.

\textbf{Learning from Feedback}: 
Various efforts have tried to improve semantic parsers based on feedback or execution validation signals. For example, ~\newcite{Clarke2010DrivingSP} and ~\newcite{artzi-zettlemoyer-2013-weakly} show that semantic parsers can be improved by learning from binary correct/incorrect feedback signals or validation functions. 
\newcite{iyer-etal-2017-learning} improve text-to-SQL parsing
by counting on humans to assess the correctness of the 
execution results generated by the inferred parses. 
In their system, parses with correct results are used
to augment the training set together with crowdsourced
gold parses of the parses that are marked as incorrect. 
\citet{lawrence-riezler-2018-improving} show that a text-to-Overpass parser can be improved using historic logs of token-level binary feedback (collected using a graphical user interface that maps an Overpass query to predefined blocks) on generated parses.
We note that our work is different from this line of work in that we do not seek to retrain and generally improve the parser, rather we focus on the task of immediately incorporating the natural language feedback to correct an initial parse. 

\textbf{Interactive Semantic Parsing} Multiple other efforts sought to
interactively involve humans in the parsing process itself.
\newcite{he2016human} ask simplified
questions about uncertain dependencies in CCG parses and use the answers as soft constraints to regenerate the parse. 
Both \newcite{li2014constructing}  and~\newcite{Su:2018:NLI:3209978.3210013} 
generate semantic parses
and present them in a graphical user interface that humans can control to edit the initial parse.
\newcite{gur-etal-2018-dialsql}
ask specific predefined multiple choice questions about a narrow set of predefined parsing errors. This interaction model together with
the synthetically generated erroneous parses that are
used for training can 
be appropriate for simple text-to-SQL parsing instance as in WikiSQL, which was the only dataset used for evaluation.
~\newcite{yao-etal-2019-model} ask yes/no
questions about the presence of SQL components  while
generating a SQL parse one component at a time. Our work falls under the general category of interactive semantic parsing. However, 
our interaction model is solely based on
natural language feedback which can convey richer
information and offering a more flexible interaction.
Our work is closest to~\cite{labutov-etal-2018-learning}, 
which also studies correcting semantic parses with 
natural language feedback, but we 
(1) focus on text-to-SQL parsing and build on 
a multi-domain dataset that requires generating
complex semantic structures and generalizing to unseen
domains (Labutov et al. consider only
the domain of email and biographical research); (2)
pair the mispredicted parses and feedback with
gold parses\footnote{In real world
scenarios, the gold parse is the final parse that the
user approves after a round (or more) of corrections.} in both our training and testing splits which
benefits a wider class of correction models; and (3)
show that incorporating the mispredicted
parse significantly improves
the correction accuracy (on the contrary to the findings of Labutov et al.).

\textbf{Asking Clarifying Questions:} Another relevant research direction focused on extending semantic parsers beyond one-shot interactions by creating agents that can ask clarifying questions that resolve ambiguities with the
original question.
For example, ~\newcite{DBLP:journals/corr/abs-1808-06740} showed that using reinforcement learning based agents that can ask clarifying questions can improve the performance of semantic parsers in the ``If-Then recipes'' domain. 
Generating clarifying questions have been studied in multiple domains to resolve ambiguity caused by speech recognition failure~\cite{Stoyanchev2014TowardsNC}, recover missing information in question answering~\cite{rao-daume-iii-2018-learning} or clarify information needs in open-domain information-seeking~\cite{Aliannejadi:2019:ACQ:3331184.3331265}. 
Our work is different from this research in that we focus on enabling and leveraging human feedback that corrects an initial parse of a
fully specified question rather than spotting and clarifying 
ambiguities.

\section{Experiments\label{sec:exprs}}
We present and evaluate a set of
baseline models for the correction
task (Section~\ref{sec:task})~
in which we use~\name~for training
and testing (unless otherwise stated).
Our main evaluation measure is the
correction accuracy---the percentage
of the testing set examples that
are corrected---in which we follow~\newcite{yu-etal-2018-spider}
and compare the corrected parse to the gold parse
using exact set match.\footnote{Exact set match is a binary measure
of exact string matching between SQL queries that handles
ordering issues.}
We also report the end-to-end accuracy on Spider development 
set (which we use to construct our testing set) of the two
turn interaction scenario: first Seq2Struct attempts to parse
the input question. If it produced a wrong parse 
the question together with that parse and the corresponding
feedback are attempted using the correction model. An example
is considered ``correct'' if any of the two attempts produces
the correct parse.\footnote{
Seq2Struct produces correct parses
for 427/1034 of Spider Dev. 
511 of the remaining examples
are supported by our SQL explanation
patterns. We estimate the end-to-end
accuracy as $(427+511*X/100)/1034$, where
$X$ is the correction accuracy.}

\begin{table*}[th!]
    \centering
    \begin{tabular}{l c c}
          \toprule
          & \multicolumn{2}{c}{\textbf{Exact Match Accuracy~(\%)}}\\
    \textbf{Baseline }& \textbf{Correction}& \textbf{End-to-End}\\
        \hline
        Without Feedback & & \\
        \ \ \ \ \ $\Rightarrow$ Seq2Struct & N/A & 41.30 \\
        \ \ \ \ \ $\Rightarrow$ Re-ranking: Uniform & 2.39 & 42.48 \\
        \ \ \ \ \ $\Rightarrow$ Re-ranking: Parser score & 11.26 & 46.86 \\
        \ \ \ \ \ $\Rightarrow$ Re-ranking: Second Best & 11.85 & 47.15 \\
        With Feedback & & \\
        \ \ \ \ \ $\Rightarrow$ Re-ranking: Handcrafted & 16.63 & 49.51 \\
        \ \ \ \ \ $\Rightarrow$ Seq2Struct+Feedback & 13.72 & 48.08 \\
        \ \ \ \ \ $\Rightarrow$ EditSQL+Feedback & \textbf{25.16} & \textbf{53.73} \\
        \midrule
        Re-ranking Upper Bound & 36.38 & 59.27 \\
        Estimated Human Accuracy & 81.50 & 81.57\\ 
                \bottomrule
    \end{tabular}
    \caption{Correction and End-to-End accuracies of Baseline models.}
    \label{tab:basline_results}
\end{table*}

\subsection{Baselines}
\textbf{Methods that ignore the feedback:} 
One approach for parse correction is re-ranking the decoder beam (top-$n$ predictions)~\cite{yin19acl}. Here, we simply discard the
top-1 candidate and
sample uniformly and with probabilities proportional
to the parser score of each candidate. We also report the
accuracy of deterministically choosing the second candidate.

\textbf{Handcrafted re-ranking with feedback:}
By definition, the feedback $f$ describes how to edit the initial 
parse $p^\prime$ to reach the correct parse. 
We represent the ``diff'' between  $p^\prime$
and each candidate parse in the beam $p_i$ as set
of schema items that appear only in one of them.
For example, the diff between \texttt{select first\_name, last\_name from students}
and \texttt{select first\_name from teachers} is \{last\_name, students, teachers\}.
We assign a score to $p_i$ 
equals to the number of matched schema items in the diff with $f$.
A schema item (e.g., ``first\_name'') is considered to be mentioned in $f$ is all the individual tokens (``first'' and ``name'') are tokens
in $f$. 

\textbf{Seq2Struct+Feedback:}
The Seq2Struct model we use to generate erroneous parses for data collection (Section~\ref{sec:generating-pairs}) reached an accuracy of 41.3\%
on Spider's development set when trained on the full 
Spider training set for 40,000 steps. After that initial training phase, 
we adapt the model to incorporating the feedback by appending
the feedback to the question for each training example in~\name~and
we continue training the model to predict the gold parse
for another 40,000 steps. We note that Seq2Struct+Feedback
does not use the mispredicted parses.

\textbf{EditSQL+Feedback:}
EditSQL~\cite{zhang19} is the current state-of-the-art
model for conversational text-to-SQL. It generates a parse
for an utterance at a conversation turn $i$ 
by editing (i.e., copying from) the parse generated
at turn $i-1$ while condition on all previous utterances
as well as the schema.
We adapt EditSQL for the correction task by providing
the question and the feedback as the utterances
at turn one and two respectively, and we force it to
use the mispredicted parse the the prediction of turn 
one~(rather than predicting it). We train the model on
the combination of the training sets of~\name~and
Spider~(which is viewed as single turn
conversations).\footnote{We exclude turn one predictions
from the training loss when processing~\name~examples otherwise,
the model would be optimized to produce the mispredicted parses.
We use the default hyper-parameters
provided by the authors together with the development set 
of~\name~for early stopping.}

To provide an estimate of \textbf{human performance}, we report the percentage of feedback instances labeled as \textit{Complete} as described in Section~\ref{sec:feedback_characteristics}. We also report
the \textbf{re-ranking upper bound}~(the percentage of test examples
whose gold parses exist in Seq2Struct beam).

\subsection{Main Results}
The results in Table~\ref{tab:basline_results} suggest
that: (1) the feedback we collect is indeed useful 
for correcting erroneous parses; (2) incorporating the
mispredicted parse helps the correction process~(even
a simple handcrafted baseline that uses the
mispredicted parases outperforms a strong trained neural model);
and (3) the state-of-the-art EditSQL model equipped with 
BERT~\cite{devlin2019bert} achieves the best performance, yet it still struggles with the task we introduce, leaving a large gap for improvement.

\subsection{Analysis\label{sec:modelanalysis}}
\textbf{Does EditSQL+Feedback use the feedback?}
To confirm that EditSQL+Feedback does not learn to
ignore the feedback, we create a test set of random feedback
by shuffling the feedback of~\name~test examples.
The accuracy on the randomized test set drops to 5.6\%.

\textbf{Is~\name~ just another conversational text-to-SQL dataset?}
We evaluate the trained EditSQL models on SParC and CoSQL
(state-of-the-art models trained by EditSQL authors) on~\name~test set, and we get
accuracies of 3.4\% and 3.2\%, respectively.
That confirms that~\name~targets different modeling aspects
as we discuss in Section~\ref{sec:related_work}.

\textbf{Is~\name~only useful for correcting Seq2Struct errors?}
EditSQL is also shown to achieve strong results on Spider~(57.6\% on 
the development set)
when used in a single-turn mode (state-of-the-art
when we started writing this paper). We collect
feedback for a sample of 179 mispredicted parses produces by
EditSQL.\footnote{We started with 200, but 21 of them turned
out to have alternative correct parses~(false negatives).} Using the EditSQL+Feedback model trained on~\name~
we get a correction accuracy of 14.6\% for EditSQL errors.

\section{Conclusions and Future Work\label{sec:conclusion}}
We introduce the task of SQL parse correction using natural language feedback
together with a dataset of human-authored feedback
paired with mispredicted and gold parses. We compare
baseline models and show that natural language
feedback is effective for correcting parses, but still
state-of-the-art models struggle to solve the task.
Future work can explore improving the correction models,
leveraging logs of natural language feedback to improve text-to-SQL
parsers, and expanding the dataset to include multiple turns of correction.

\section*{Acknowledgments}
 We thank our ACL reviewers for their feedback and suggestions.
 Ahmed Elgohary completed part of this work
 while being supported by a grant from
 the Defense Advanced Research Projects Agency and Air Force Research Laboratory, and awarded to Raytheon BBN Technologies under contract number FA865018-C-7885.

\bibliography{journal-full,ahmed,acl2020}
\bibliographystyle{style/acl_natbib}

\appendix
\section{Appendix\label{sec:appendix}}
\begin{figure*}[t!]
\begin{framed}
\footnotesize
\textbf{Correcting Steps for Answering Questions.}
\begin{enumerate}
\item We have some information stored in tables; each row is a record that consists of one or more columns. Using the given tables, we can answer questions by doing simple systematic processing steps over the tables. Notice that the answer to the question is always the result of the last step. Also, notice that the steps might not be in perfect English as they were generated automatically. Each step, generates a table of some form.

\item For each question, we have generated steps to answer it, but it turned out that something is wrong with the steps. \textbf{You task} is write down in English a short (one sentence most of the time) description of the mistakes and how it can be correct.  It is important to note that we are not looking for rewritings of steps, but rather we want to get short natural English commands (15 words at most) that describes the correction to be made to get the correct answer. 

\item Use proper and fluent English. Don't use math symbols.

\item Don't rewrite the steps after correcting them. But rather, just describe briefly the change that needs to be made.

\item We show only two example values from each table to help you understand the contents of each table. Tables typically contain several rows. Never use the shown values to write your input.

\item There could be more than one wrong pieces in the steps. Please, make sure to mention all of them not just one.

\item If the steps are correct, just check the ``All steps are correct'' box

\item Some of the mistakes are due to additional steps or parts of steps. Your feedback can suggest removing those parts.

\item Do not just copy parts of the questions. Be precise in your input.

\item If in doubt about how to correct a mistake, just mention what is wrong.

\item You do not have to mention which steps contain mistakes. If in doubt, do not refer to a particular step.

\item The generated steps might not sound like the smartest way for answering the question. But it is the most systematic way that works for all kinds of questions and all kinds of tables. Please, do not try to propose smarter steps.

\end{enumerate}
\end{framed}
\caption{Crowd-sourcing instructions\label{sql_instr}}
\end{figure*}

\subsection{Feedback Collection instructions}

Figure~\ref{sql_instr} shows the instructions shown to the annotators.

\subsection{Feedback Collection Interface Screenshot}

\begin{figure*}[!b]
  \centering
  \includegraphics{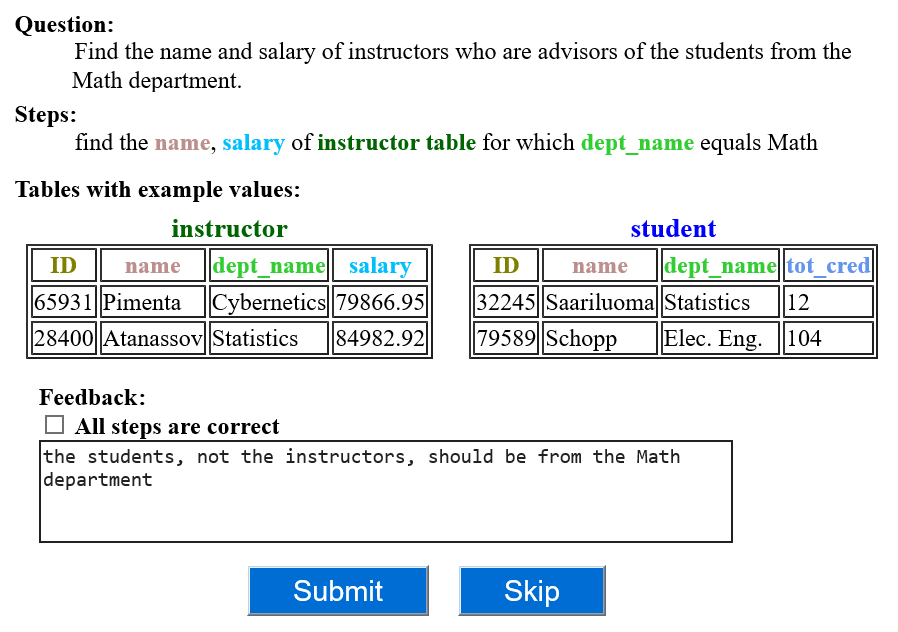}
    \caption{An example of the data collection interface. The Predicted SQL is: \texttt{'SELECT name, salary FROM instructor WHERE dept\_name LIKE "\%math\%"'}. Note that neither the gold nor the predicted SQL are shown to the annotator.}
  \label{fig:interface_example}
 \end{figure*}

 Figure~\ref{fig:interface_example} shows an example of the data collection interface. The Predicted SQL is: \texttt{'SELECT name, salary FROM instructor WHERE dept\_name LIKE "\%math\%"'}. Note that neither the gold nor the predicted SQL are shown to the annotator.

\subsection{Example of Explanations}
Figure~\ref{fig:explanation_patterns} shows several examples of how different SQL components can be explained in natural language.

\begin{figure*}
\begin{framed}
\footnotesize

\begin{tabular}{p{3cm} p{10cm}}
\textbf{SQL Component} &	\textbf{Explanation} \\ \midrule
intersect & show the rows that are in both the results of step 1 and step 2 \\
union & show the rows that are in any of the results of step 1 and step 2 \\
except & show the rows that are in the results of step 1 but not in the results of step 2 \\
limit n & only keep the first  n rows of the results in step 1 \\
join & for each row in Table 1, find corresponding rows in Table 2 \\
select & find Column of Table \\
aggregation & find each value of Column1 in Table along with the OPERATION of Column2 of the corresponding rows to each value \\
ordering & order Direction by Column \\
condition & whose Column Operation Value \\
distinct & without repetition \\
  
\end{tabular}
\end{framed}
\caption{Examples of how different SQL components can be explained in natural language}
\label{fig:explanation_patterns}

\end{figure*}

\end{document}